\documentclass[review,12pt]{elsarticle}

\usepackage{amssymb}
\usepackage{amsmath}
\usepackage{booktabs}
\usepackage{pgfplots}
\pgfplotsset{compat=1.18}
\pgfplotsset{
    label style={font=\normalsize},
    tick label style={font=\normalsize},
    legend style={font=\normalsize}
}
\usepackage{float}
\usepackage{placeins}
\usepackage{caption}
\usepackage{makecell}
\usepackage{xcolor}

\usepackage{tikz}
\usepackage{subcaption}
\usepackage{multirow}
\usetikzlibrary{shapes.geometric, arrows, positioning, calc, patterns}
\usepgfplotslibrary{statistics}
\usepackage{microtype}

\usepackage{url}

\journal{Engineering Applications of Artificial Intelligence}

\begin{document}
\begin{frontmatter}

\title{Vision-Language Models for Analog Gauge Reading: An Empirical Study of Specialization, Transfer and Reliability}

\author[1]{Abdul Mueez}
\author[1]{Aaditya Baranwal}
\author[1]{Junior Chaj-Mejia}
\author[2]{Guneet Bhatia}
\author[2]{Jason T. Voelker}
\author[1]{Shruti Vyas\corref{cor1}}

\cortext[cor1]{Corresponding author. Email: shruti@ucf.edu. Full address: 4356 Scorpius St., Orlando, FL 32816, United States of America.}

\address[1]{University of Central Florida, Orlando, FL, United States}
\address[2]{Siemens Energy, Orlando, FL, United States}

\begin{abstract}
Analog gauges remain common in industrial environments where manual inspection is costly or hazardous. The engineering application addressed here is direct numerical reading of single-target analog-gauge images, while the artificial-intelligence contribution is a systematic evaluation of specialization, transfer, robustness and reliability for a general-purpose vision-language model (VLM) without an explicit pointer-segmentation and geometric-reading pipeline. The Qwen2.5-VL-7B-Instruct model is evaluated using zero-shot prompting, in-context learning (ICL) and parameter-efficient fine-tuning with Quantized Low-Rank Adaptation (QLoRA) on a public synthetic dataset, a video-derived Pressure Gauge dataset and a proprietary industrial dataset. All fine-tuning experiments use a fixed 20-epoch protocol with the final epoch used for analysis; separate models with and without supplied gauge ranges remove prompt-setting confounds. The primary metric is range-normalized mean percentage error (MPE). The best fine-tuned MPE values are 2.39\% on the synthetic dataset, with a 95\% bootstrap confidence interval (CI) of 1.43--3.90\%; 2.61\% on the Pressure Gauge dataset, with a CI of 1.66--3.80\%; and 4.43\% on the proprietary industrial dataset, with a CI of 2.31--7.14\%. Leave-one-dataset-out experiments reveal substantial transfer degradation on held-out synthetic and proprietary data, while robustness tests identify Gaussian blur as the strongest tested corruption. Reliability analysis shows that high-confidence errors remain possible, motivating abstention and independent validation in safety-critical use. These results support QLoRA-specialized VLMs for direct single-gauge reading but not yet a deployment-ready plant-monitoring pipeline.
\end{abstract}




\begin{keyword}
Industrial automation
\sep Analog gauge reading
\sep Vision-language model
\sep Quantized low-rank adaptation
\sep Parameter-efficient fine-tuning
\end{keyword}

\end{frontmatter}

\section{Introduction}
\label{sec:introduction}

Analog gauges are ubiquitous instruments for monitoring critical parameters such as pressure, temperature and flow rate across manufacturing facilities, power substations, process plants and other industrial environments \citep{YangEtAl2025Info, TrairattanapaEtAl2022Isai, LiEtAl2020JPhysConf, TianEtAl2024Mipr, HowellsEtAl2021CVPRW}. Accurate and timely readings are important for operational safety, maintenance and process oversight, yet repeated manual inspection is labor intensive, vulnerable to transcription error and potentially hazardous in difficult-to-access locations \citep{YangEtAl2025Info, TianEtAl2024Mipr}.

Analog gauges persist because they are simple, locally readable and can operate without an electrical data interface. Their mechanical simplicity can also be attractive in harsh or electromagnetically noisy environments \citep{kappert2022sensor}. Rather than replacing this installed base, an image-based monitoring layer can digitize readings while preserving the underlying instrument.

Traditional automated gauge readers commonly decompose the problem into gauge localization, perspective correction, pointer or keypoint detection, scale extraction and geometric conversion \citep{PeixotoEtAl2022Telecom, HowellsEtAl2021CVPRW, ReitsmaEtAl2024Icra, YangEtAl2025Info}. These pipelines can be effective, but each stage introduces assumptions about geometry, image quality or gauge design. Direct regression approaches reduce some of this complexity, although data scarcity and domain shift remain important concerns \citep{TrairattanapaEtAl2022Isai, NinamaEtAl2024SciRep, ji2024another}.

Large vision-language models (VLMs) provide a different starting point: a pretrained multimodal representation can be adapted to produce the gauge value directly from an image and textual instruction. However, strong generic visual-language capability does not imply metrological reliability. The central objective of this work is therefore to determine how much task-specific specialization is required for accurate analog-gauge reading and to characterize where a specialized VLM remains unreliable.


This study makes four contributions:
\begin{enumerate}
    \item It systematically compares zero-shot prompting, in-context learning (ICL) and QLoRA (Quantized Low-Rank Adaptation)-based parameter-efficient fine-tuning of Qwen2.5-VL-7B-Instruct.
    
    \item It trains separate with-range and without-range models, providing a clean metadata ablation.
    
    \item It adds leave-one-dataset-out (LODO) and adverse-condition corruption analyses to probe transfer and robustness.
    
    \item It expands evaluation beyond MPE through bootstrap confidence intervals, MAE, RMSE, tail-error statistics, operational tolerance rates, parsing failures, calibration metrics and explanation-method comparisons.
\end{enumerate}

The scope is deliberately narrower than full plant-scene automation. Each evaluated input contains one target gauge, but it is not necessarily a tight gauge crop: multiple images retain substantial surrounding scene context and the gauge occupies only a small portion of the frame. Only source images containing multiple gauges in one of the datasets were cropped to isolate one gauge per sample. The model predicts the reading directly from these single-target-gauge images; detecting and selecting among multiple gauges and plant-level acquisition or alarm integration remain outside the evaluated pipeline.

\section{Related Work}
\label{sec:related_work}

\subsection{Multi-stage computer-vision gauge readers}
Classical analog-gauge readers typically use geometric image processing. Circular boundaries can be estimated with Hough transforms or ellipse fitting, the pointer can be extracted with line detection, polar projection or principal-component analysis and the final value can be computed from the pointer angle and scale geometry \citep{GellaboinaEtAl2013, LauridsenEtAl2019Visigrapp, LiEtAl2019Geometric, PeixotoEtAl2022Telecom}. These methods are interpretable and computationally efficient, but sensitivity to glare, viewing angle, occlusion and non-standard dial layouts can propagate through the pipeline.

Recent learning-based systems retain much of this staged structure while replacing individual components with neural networks. Examples include segmentation-based pointer-meter reading \citep{ZuoEtAl2020}, detection and recognition systems for substations \citep{LiuEtAl2020Measurement}, keypoint-based mobile gauge transcription \citep{HowellsEtAl2021CVPRW}, GaugeTracker \citep{TianEtAl2024Mipr} and learning-based gauge reading in unconstrained imagery \citep{ReitsmaEtAl2024Icra}. A recent outdoor-substation method combines learned dial segmentation with distortion correction and scale recognition \citep{YangEtAl2025Info}. Collectively, these approaches show that strong performance is possible, but they also illustrate the diversity of preprocessing, geometric assumptions and supervision required across gauge-reading systems.

\subsection{Direct regression and compact learned baselines}
An alternative is to regress the reading directly, avoiding explicit pointer geometry. CNN-based direct regressors and multi-task regression systems have demonstrated the feasibility of this formulation \citep{TrairattanapaEtAl2022Isai, NinamaEtAl2024SciRep, ji2024another}. MobileNetV2 \citep{sandler2018mobilenetv2} is an attractive compact backbone because of its low computational cost. However, direct regression can become difficult when the same apparent pointer position corresponds to different engineering values across gauge ranges and evaluation can be confounded when different target parameterizations are used across experimental conditions. For this reason, the MobileNetV2 comparison in this work is treated as a lightweight reference rather than a clean test of whether range metadata is visually inferred.

\subsection{Vision-language models and model selection}
VLMs couple visual representations with autoregressive language generation, providing a flexible interface for tasks that combine visual parsing and textual instructions \citep{bordes2024introduction}. Qwen2.5-VL-7B-Instruct was selected because it is an open-weight model with documented dynamic-resolution visual processing and strong general-purpose localization, document and structured visual-understanding capability \citep{Qwen25VL}. These properties are relevant to analog gauges, where small ticks, printed scale values and pointer geometry must be interpreted jointly. The 7B-scale model is also small enough to adapt locally with QLoRA while retaining substantially greater capacity than compact VLM families such as SmolVLM2.

The present study does not claim an exhaustive same-split comparison against every contemporary industrial inspection architecture. Instead, it asks whether specialization can turn a general-purpose VLM into an accurate direct gauge reader and then subjects that model to controlled metadata ablation, cross-domain transfer, robustness and reliability analyses. This distinction is important because published gauge-reading systems often use different datasets, crop policies, sampling strategies and unavailable implementations, preventing a fully controlled cross-paper ranking.

\section{Datasets}
\label{sec:datasets}

This work is evaluated on three datasets: the Synthetic dataset, the Pressure Gauge video dataset and the Siemens Energy (SE) dataset. Descriptive dataset names are used throughout the paper for readability. For compact labels in tables and figures, \textit{Pressure} denotes the Pressure Gauge dataset.

\subsection{Synthetic Dataset}
\begin{figure}[!htbp]
    \centering
    \includegraphics[width=\linewidth]{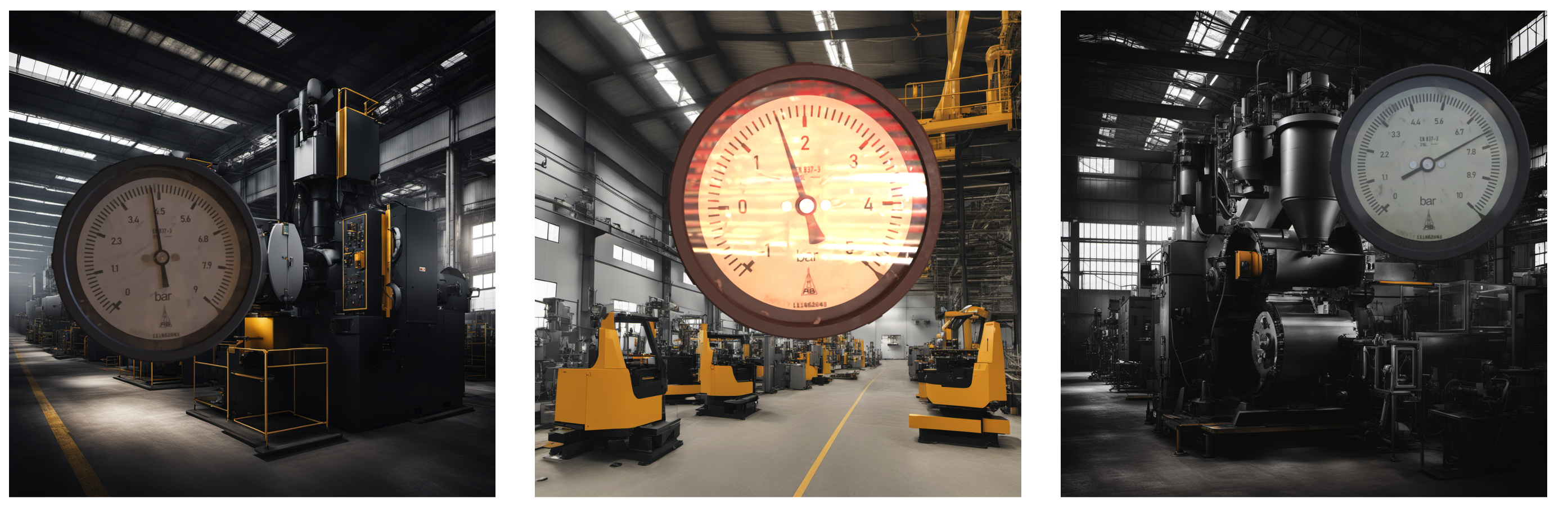}
    \caption{Examples from the public Synthetic Data for Precision Gauge Reading dataset \citep{EndavaKaggle}.}
    \label{synth_fig}
\end{figure}

A synthetic dataset titled ``Synthetic Data for Precision Gauge Reading'' created by Endava and distributed through Kaggle \citep{EndavaKaggle} was used. The ``Synthetic DS6.0'' subset used here contains 501 gauge images with annotations for the reading and scale information. A random train-test split produced 375 training images and 126 test images. The source dataset contains additional component masks and keypoints, but only the gauge reading and scale endpoints are required by the present experiments. Representative Synthetic samples are shown in Figure~\ref{synth_fig}.

\subsection{Pressure Gauge Video Dataset}
The Pressure Gauge dataset is derived from the ``Pressure Gauge Reader Data'' video collection \citep{grassme2022pressure, LauridsenEtAl2019Visigrapp}. Frames were extracted at one-second intervals and then filtered to retain unique needle positions. The source training partition consists of three videos of the same gauge captured under slightly different lighting and framing conditions. The source test partition contains seven videos: two show the same gauge with minor lighting/framing changes, two introduce angle variation and three feature different gauges. The final split contains 57 training images and 36 test images. Importantly, the source dataset separates training and test videos: the training images used here come from videos \texttt{out2}, \texttt{out3} and \texttt{out5}, whereas the test images come from \texttt{man1} through \texttt{man7}. Consequently, no video source is shared between the Pressure Gauge training and test partitions. This establishes video-level separation, but does not by itself establish gauge-identity or acquisition-condition independence. The exact filename-level split will be included in the public release described in Section~\ref{sec:data_code}.

\begin{figure}[!htbp]
    \centering
    \includegraphics[width=\linewidth]{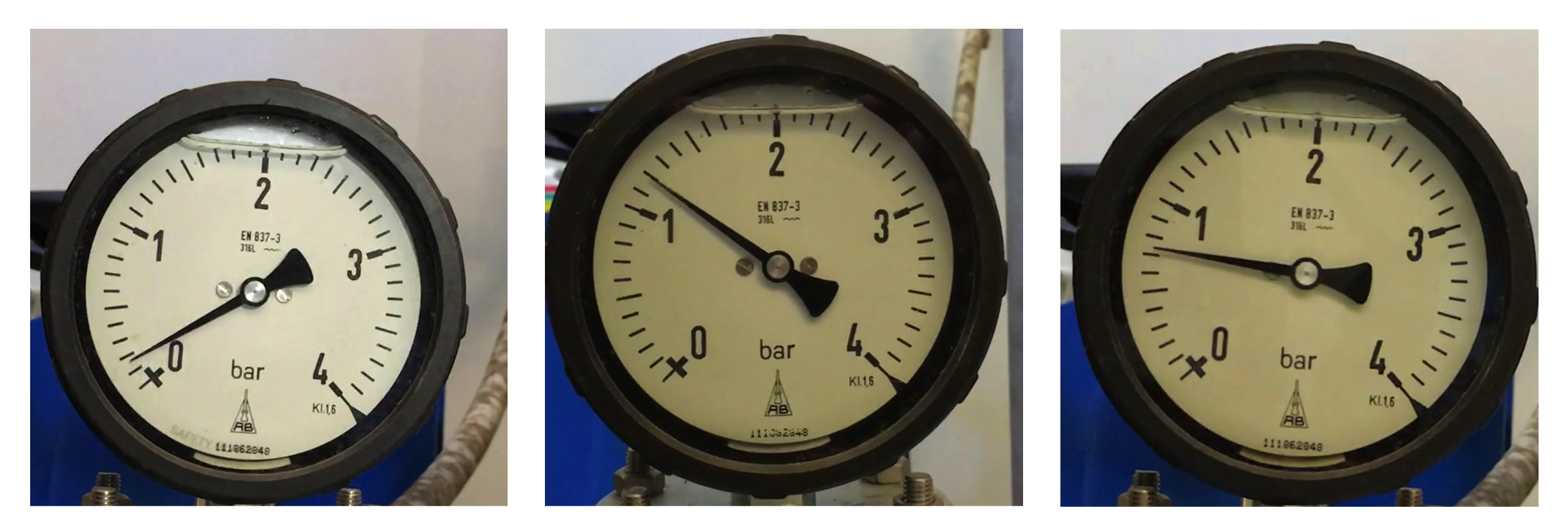}
    \caption{Example frames from the Pressure Gauge video collection \citep{grassme2022pressure}.}
    \label{k_p_dataset}
\end{figure}

The source dataset does not provide the numerical annotations required by the present regression evaluation. A first annotator assigned the gauge reading and scale endpoints for each selected frame. A second person then verified all annotations and made changes as necessary. Representative Pressure Gauge frames are shown in Figure~\ref{k_p_dataset}.

\subsection{Siemens Energy Dataset}
The SE dataset contains gauge photographs acquired in real industrial environments. Raw images cannot be publicly distributed because of proprietary and site-sensitivity constraints. The image collection was randomly divided into approximately 75\% training and 25\% testing, yielding 132 original training images and 43 test images. Only training images containing more than one gauge were cropped to isolate individual gauge instances, increasing the training count to 152. Images that already contained one gauge were retained without imposing a tight gauge crop; consequently, some samples include substantial surrounding scene context and the target gauge occupies only a small portion of the image. The 43 test images each contain one target gauge and required no multi-gauge instance extraction.

Each SE sample used for training or evaluation is associated with the numerical reading and minimum/maximum scale values. A first annotator assigned the reading and scale endpoints, after which a second person verified all annotations and made changes as necessary. To improve reproducibility without releasing sensitive imagery, the final adapters, exact split manifests, gauge-range metadata, ground-truth readings and evaluation scripts will be released.

\begin{figure}[!htbp]
    \centering
    \includegraphics[width=\linewidth]{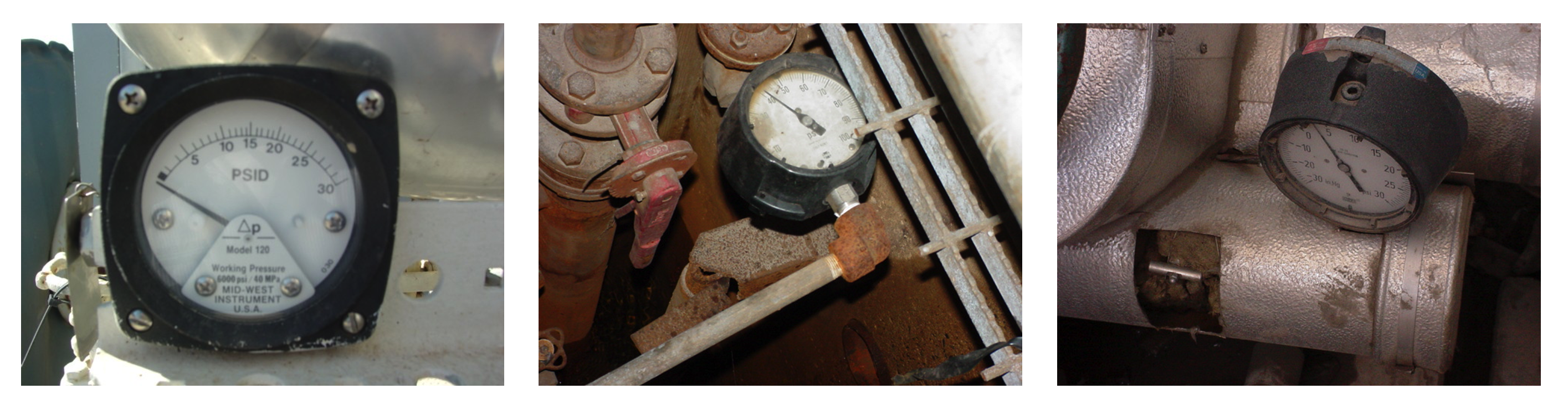}
    \caption{Examples from the proprietary SE dataset, illustrating variation in gauge appearance, background, scale and viewpoint.}
    \label{se_dataset}
\end{figure}

The SE split is random rather than source or gauge-family-held-out. Therefore, visually related gauge types or acquisition environments may occur in both partitions. The LODO experiment that excludes all SE images from fine-tuning provides complementary evidence about transfer to the SE domain, but it cannot determine whether particular SE gauge types overlap between the random training and test partitions. A validated gauge-family taxonomy and a future gauge-family or plant-held-out split are required for that question. Representative SE images are shown in Figure~\ref{se_dataset}.

\subsection{Dataset Summary}
Table~\ref{tab:dataset_summary} summarizes the final dataset split sizes, the number of distinct scale configurations, overall scale endpoints and test-set ground-truth value ranges.

\begin{table}[!htbp]
\centering
\caption{Summary of the final dataset splits, with scale-range and value statistics. ``\# Ranges'' is the number of distinct (min, max) scale configurations present in the split; ``Value Range'' is the span of ground-truth needle readings.}
\label{tab:dataset_summary}
\scriptsize
\setlength{\tabcolsep}{1pt}
\begin{tabular*}{\linewidth}{@{\extracolsep{\fill}}p{0.18\linewidth}cccp{0.18\linewidth}p{0.16\linewidth}@{}}
\toprule
\textbf{Dataset} & \makecell{\textbf{Train}} & \makecell{\textbf{Test}} & \makecell{\textbf{\# Ranges}\\\textbf{(train/test)}} & \makecell{\textbf{Scale Endpoints}\\\textbf{(min--max)}} & \makecell{\textbf{Value Range}\\\textbf{(test)}} \\
\midrule
Synthetic & 375 & 126 & 7 / 7 & $-2$ to $10$ & $-1.9$ to $9.0$ \\
Pressure & 57 & 36 & 1 / 3 & $-1$ to $6$ & $-0.42$ to $3.2$ \\
SE (Ours) & 152 & 43 & 32 / 18 & $-30$ to $5000$ & $0$ to $2300$ \\
\bottomrule
\end{tabular*}
\end{table}

\section{Methodology}
\label{sec:methodology}

\subsection{Qwen2.5-VL-7B-Instruct}
\begin{figure}[!htbp]
    \centering
    \includegraphics[width=0.7\linewidth]{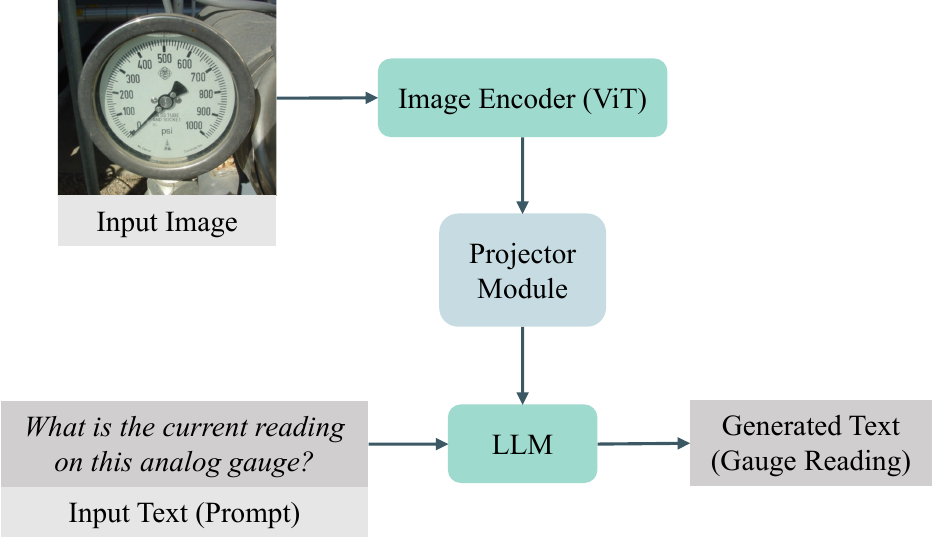}
    \caption{General VLM structure: a vision encoder extracts visual tokens, a projector/merger maps them to the language-model representation and the autoregressive language model produces the numerical answer.}
    \label{fig:vlm_schematic}
\end{figure}

The core model is \texttt{Qwen/Qwen2.5-VL-7B-Instruct} \citep{Qwen25VL}. A parameter audit of the final checkpoint after 20 training epochs reports 8,315,961,344 parameters in total: 7,615,616,512 in the language-model component, 631,975,680 in the vision encoder, 44,574,464 in the visual projector/merger and 23,794,688 LoRA parameters. The LoRA parameters therefore represent 0.29\% of the loaded parameter count. The checkpoint is loaded for inference with parameters frozen; the LoRA count is identified from the adapter parameter names rather than from \texttt{requires\_grad} at inference time. Figure~\ref{fig:vlm_schematic} illustrates the vision-encoder, projector/merger and language-model structure used in the VLM formulation.

\subsection{Zero-shot and in-context learning}
For zero-shot evaluation, the model receives the gauge image and the task instruction without demonstrations. ICL prepends nested sets of training examples. For each dataset, the 50-example pool was selected by random sampling from the corresponding training partition and the smaller example sets were nested subsets of that sampled pool. Thus, example sets of size $k \in \{1,5,10,20,50\}$ satisfy $E_{1}\subset E_{5}\subset E_{10}\subset E_{20}\subset E_{50}$, allowing sensitivity to context size to be inspected without changing the smaller sets when additional examples are added.

\subsection{QLoRA specialization and fixed-epoch training protocol}
QLoRA \citep{QLoRA} combines low-rank adaptation \citep{LoRA} with quantized frozen base weights. During training, the Qwen base model is loaded with 4-bit NF4 quantization, double quantization and bfloat16 compute. LoRA uses rank $r=8$, $\alpha=16$ and dropout 0.05 on \texttt{q\_proj}, \texttt{k\_proj}, \texttt{v\_proj}, \texttt{o\_proj}, \texttt{gate\_proj}, \texttt{up\_proj} and \texttt{down\_proj} modules.

Five Qwen2.5-VL-7B models are trained: an all-domain model trained and evaluated with range, an all-domain model trained and evaluated without range and three LODO models in which Synthetic, Pressure Gauge or SE is excluded from training and then used only for testing. The LODO models use the with-range prompt so that the held-out-domain comparison changes training-domain coverage rather than prompt formulation.

The reported protocol uses no separate validation set and no early stopping. All five revised fine-tuning runs were restarted from the base Qwen2.5-VL-7B-Instruct checkpoint, trained for 20 fixed epochs and the final checkpoint after epoch 20 is used for every reported analysis. No checkpoint is selected using test performance.

Training uses a batch size of 1 with gradient accumulation over 8 steps, AdamW with a learning rate of $2\times10^{-4}$, bfloat16 mixed precision and two Nvidia RTX A5500 GPUs. Training images are augmented with random rotation ($\pm15^{\circ}$, probability 0.7), color jitter (brightness 0.3, contrast 0.3, saturation 0.2, hue 0.1; probability 0.8), Gaussian blur (kernel 3, $\sigma=0.1$--1.5; probability 0.3) and random perspective transformation (distortion scale 0.2, probability 0.5, bilinear interpolation). Test images are not augmented. The Qwen processor uses dynamic image resolution with total-pixel bounds \texttt{256*28*28} to \texttt{640*28*28}. A separate audit was not performed to verify label validity for random-crop outcomes; accordingly, crop-related label preservation is not claimed as an independently validated property of the augmentation pipeline. 

\subsection{Prompt templates}
The reported experiments use the templates in Table~\ref{tab:prompts}. Preliminary prompt variants were not systematically archived and are therefore not presented as experimental results.

\begin{table}[!htbp]
\centering
\caption{Prompt templates used in the reported experiments. The same range condition is used during training and inference for the fine-tuned models.}
\label{tab:prompts}
\small
\begin{tabular}{@{}ll p{0.55\textwidth}@{}}
\toprule
\textbf{Task} & \textbf{Role / Condition} & \textbf{Template} \\
\midrule
\multirow{3}{*}{\textbf{Value Reading}} & System & {\ttfamily \raggedright You are an assistant that reads analog gauges. Respond with *only* the numerical value the needle is pointing to. For example, if the needle points to 5.5, respond `5.5'.} \\
\cmidrule(l){2-3}
& User (With range) & {\ttfamily \raggedright What is the current reading on this analog gauge? The scale ranges from \{min\_val\} to \{max\_val\}.} \\
\cmidrule(l){2-3}
& User (No range) & {\ttfamily \raggedright What is the current reading on this analog gauge?} \\
\midrule
\multirow{2}{*}{\textbf{Scale Probe}} & System & {\ttfamily \raggedright You are a vision assistant that reads analog dial gauges. Look only at the printed scale numbers on the dial face. Report the smallest and largest numbers printed on the scale. Respond ONLY as JSON: \{"min": <number>, "max": <number>\}.} \\
\cmidrule(l){2-3}
& User & {\ttfamily \raggedright What are the minimum and maximum scale values printed on this gauge's dial?} \\
\bottomrule
\end{tabular}
\end{table}

\subsection{Metrics and statistical analysis}
\label{subsec:metrics}
For sample $i$, range-normalized percentage error is
\[
 e_i=\frac{|y_i-\hat y_i|}{y_{\max,i}-y_{\min,i}}\times 100\%.
\]
The primary MPE is the mean of $e_i$. Because capping errors at 100\% can hide catastrophic failures, this study reports \textbf{both capped and uncapped MPE}. Capped MPE uses $\min(e_i,100)$; uncapped MPE retains the complete tail. Additional metrics include raw engineering-value MAE and RMSE, maximum and 95th-percentile absolute/full-scale errors, failure rate above 5\% of full scale, pass rates within $\pm1\%$, $\pm2\%$ and $\pm5\%$ of full scale and parsing-failure rate. The MAE/RMSE values are computed directly on the numerical targets and are reported as raw engineering-value errors rather than as physical-unit-specific metrology metrics.

MPE uncertainty is quantified with 10,000-sample nonparametric bootstrap 95\% confidence intervals. Paired model comparisons align predictions by filename and use both a paired bootstrap CI for the mean per-image error difference and a Wilcoxon signed-rank test.

Model output parsing first attempts strict floating-point conversion and then a numeric-expression extraction. Any remaining nonnumeric output is counted as a parsing failure rather than silently assigned a value.


\subsection{Confidence and calibration}
\label{sec:confidence_and_calibration}
For a generated numeric answer containing $m$ non-EOS output tokens, sequence confidence is defined as the geometric mean of the token probabilities assigned to the greedily generated tokens:
\[
 C=\exp\left(\frac{1}{m}\sum_{k=1}^{m}\log p_k\right).
\]
This is a token-generation score, \emph{not} a calibrated probability of numeric correctness. Calibration is therefore evaluated explicitly using 10-bin expected calibration error (ECE), maximum calibration error (MCE), reliability diagrams and risk-coverage curves with area under the risk-coverage curve (AURC). A prediction is treated as operationally correct for calibration when its absolute error is within 5\% of full scale. This 5\% threshold is an evaluation tolerance for the reliability analysis, not a claim of compliance with a metrology or instrument-accuracy standard. High-confidence (HC) errors are defined as incorrect predictions (error exceeding 5\% of full scale) with sequence confidence $C\ge0.99$.

\section{Results}
\label{sec:results}

\subsection{Zero-shot and ICL baselines}
\begin{figure}[!htbp]
    \centering
    \begin{subfigure}[b]{\textwidth}
        \centering
        \begin{tikzpicture}
            \begin{axis}[title={Synthetic},xlabel={Number of Contextual Examples},ylabel={MPE (\%)},legend pos=north east,grid=major,width=0.9\linewidth,height=0.28\linewidth,xtick={0,1,5,10,20,50}]
            \addplot[blue,thick,mark=*] coordinates {(0,21.33) (1,26.32) (5,27.73) (10,22.94) (20,18.43) (50,24.13)};
            \addlegendentry{With Range}
            \addplot[red,thick,mark=square*] coordinates {(0,18.12) (1,25.19) (5,25.29) (10,17.79) (20,15.83) (50,21.64)};
            \addlegendentry{Without Range}
            \end{axis}
        \end{tikzpicture}
    \end{subfigure}

    \begin{subfigure}[b]{\textwidth}
        \centering
        \begin{tikzpicture}
            \begin{axis}[title={Pressure},xlabel={Number of Contextual Examples},ylabel={MPE (\%)},legend pos=north east,grid=major,width=0.9\linewidth,height=0.30\linewidth,xtick={0,1,5,10,20,50}]
            \addplot[blue,thick,mark=*] coordinates {(0,25.29) (1,20.86) (5,8.34) (10,8.72) (20,13.29) (50,10.56)};
            \addlegendentry{With Range}
            \addplot[red,thick,mark=square*] coordinates {(0,15.69) (1,11.89) (5,8.13) (10,5.89) (20,8.53) (50,6.84)};
            \addlegendentry{Without Range}
            \end{axis}
        \end{tikzpicture}
    \end{subfigure}

    \begin{subfigure}[b]{\textwidth}
        \centering
        \begin{tikzpicture}
            \begin{axis}[title={SE},xlabel={Number of Contextual Examples},ylabel={MPE (\%)},legend pos=north east,grid=major,width=0.9\linewidth,height=0.30\linewidth,xtick={0,1,5,10,20,50}]
            \addplot[blue,thick,mark=*] coordinates {(0,39.12) (1,44.81) (5,39.73) (10,27.07) (20,28.22) (50,30.36)};
            \addlegendentry{With Range}
            \addplot[red,thick,mark=square*] coordinates {(0,31.06) (1,44.26) (5,32.74) (10,32.58) (20,29.87) (50,32.98)};
            \addlegendentry{Without Range}
            \end{axis}
        \end{tikzpicture}
    \end{subfigure}
    \caption{Zero-shot (0 examples) and ICL sensitivity to the number of contextual examples. Lower MPE is better.}
    \label{fig:icl_combined_results}
\end{figure}
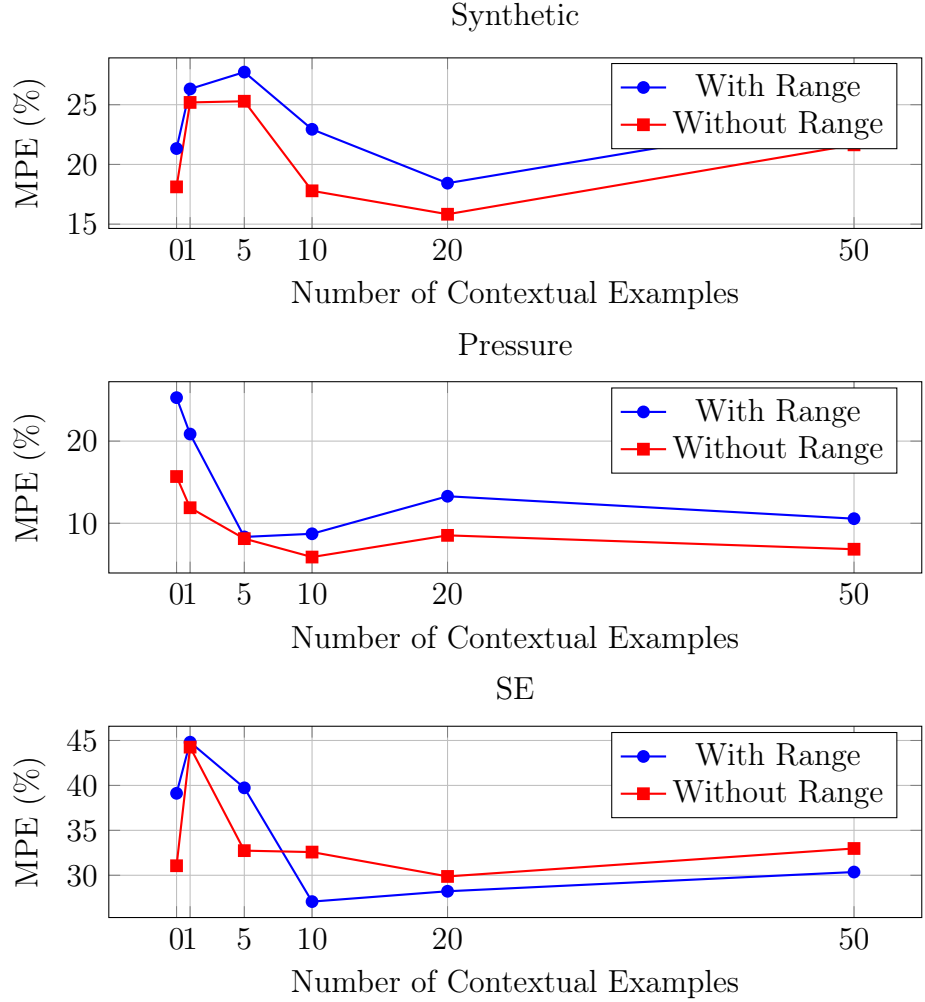

Figure~\ref{fig:icl_combined_results} shows zero-shot and ICL benchmark performance. ICL is evaluated with 1, 5, 10, 20 and 50 nested examples.

The baseline trends are unstable. The Synthetic dataset reaches its best ICL result at 20 examples without range (15.83\% MPE), Pressure Gauge at 10 examples without range (5.89\%) and SE at 10 examples with range (27.07\%). Increasing context beyond the best point does not monotonically improve accuracy, motivating weight adaptation rather than relying on context size alone.

\subsection{Fine-tuned QLoRA performance and clean range ablation}
\label{subsec:finetuned_results}
\begin{figure}[!htbp]
    \centering
    \begin{tikzpicture}
        \begin{axis}[ybar,bar width=12pt,width=\textwidth,height=0.32\textwidth,ylabel={Best MPE (\%)},symbolic x coords={Synthetic,Pressure,SE},xtick=data,ymin=0,ymax=35,enlarge x limits=0.30,legend style={at={(0.5,0.97)},anchor=north,legend columns=-1}]
        \addplot[fill=blue!55,draw=blue!80!black,bar shift=-14pt] coordinates {(Synthetic,18.12) (Pressure,15.69) (SE,31.06)};
        \addlegendentry{Zero-shot}
        \addplot[fill=orange!70,draw=orange!90!black,bar shift=0pt] coordinates {(Synthetic,15.83) (Pressure,5.89) (SE,27.07)};
        \addlegendentry{ICL}
        \addplot[fill=green!60!black,draw=green!40!black,bar shift=14pt] coordinates {(Synthetic,2.39) (Pressure,2.61) (SE,4.43)};
        \addlegendentry{Fine-tuned Qwen2.5-VL-7B}
        \end{axis}
    \end{tikzpicture}
    \caption{Best observed MPE for each learning strategy. For Pressure, the best QLoRA result is the separately trained no-range model; for Synthetic and SE, the best QLoRA result uses range metadata.}
    \label{fig:best_mpe_comparison}
\end{figure}
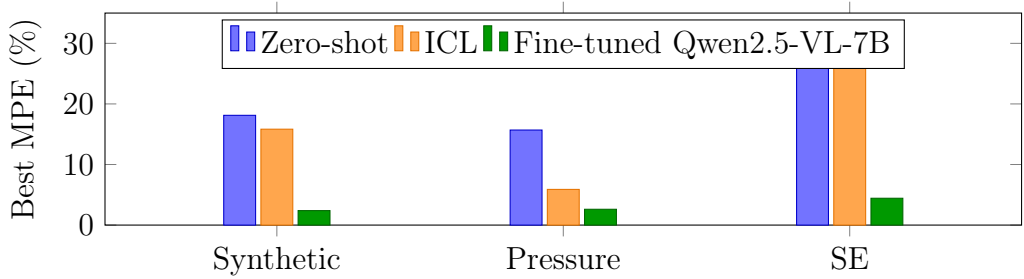

Table~\ref{tab:final_comparison} reports the fine-tuned results obtained from the final checkpoint after 20 fixed training epochs. The best MPE is 2.39\% for Synthetic with range, 2.61\% for Pressure Gauge without range and 4.43\% for SE with range. Figure~\ref{fig:best_mpe_comparison} provides a visual comparison of the best zero-shot, ICL and fine-tuned results.

\begin{table}[!htbp]
\centering
\caption{MPE (\%) for zero-shot, best-case ICL and fine-tuned Qwen2.5-VL-7B models. Fine-tuned entries include the 95\% bootstrap CI in brackets and come from separately trained with-range and without-range models; all fine-tuned results use the final checkpoint after 20 fixed epochs.}
\label{tab:final_comparison}
\small
\begin{tabular}{llcc}
\toprule
\textbf{Dataset} & \textbf{Method} & \textbf{With Range} & \textbf{Without Range} \\
\midrule
\multirow{3}{*}{Synthetic} & Zero-shot & 21.33 & 18.12 \\
 & ICL (best) & 18.43 ($k=20$) & 15.83 ($k=20$) \\
 & Fine-tuned & \makecell{\textbf{2.39}\\{[1.43, 3.90]}} & \makecell{2.73\\{[1.57, 4.58]}} \\
\midrule
\multirow{3}{*}{Pressure} & Zero-shot & 25.29 & 15.69 \\
 & ICL (best) & 8.34 ($k=5$) & 5.89 ($k=10$) \\
 & Fine-tuned & \makecell{3.31\\{[2.27, 4.57]}} & \makecell{\textbf{2.61}\\{[1.66, 3.80]}} \\
\midrule
\multirow{3}{*}{SE} & Zero-shot & 39.12 & 31.06 \\
 & ICL (best) & 27.07 ($k=10$) & 29.87 ($k=20$) \\
 & Fine-tuned & \makecell{\textbf{4.43}\\{[2.31, 7.14]}} & \makecell{6.28\\{[2.47, 11.84]}} \\
\bottomrule
\end{tabular}
\end{table}

The paired range ablation in Table~\ref{tab:range_ablation} shows that the average capped-MPE differences between the separately trained range/no-range models are not statistically significant at $p<0.05$ for any dataset. However, capped MPE alone obscures tail behavior. On SE, the no-range model has an uncapped MPE of 13.42\% compared with 4.43\% with range and its maximum full-scale error reaches 407.14\% compared with 41.25\%. Thus, range metadata is best interpreted as reducing extreme SE failures rather than as providing a statistically established mean-error improvement in this small test set.

\begin{table}[!htbp]
\centering
\caption{Paired range-ablation statistics for the separately trained fine-tuned Qwen2.5-VL-7B models. The mean difference is with-range minus without-range capped per-image percentage error; the corresponding MPE values are reported in Table~\ref{tab:final_comparison}.}
\label{tab:range_ablation}
\small
\begin{tabular}{lccc}
\toprule
\textbf{Dataset} & \makecell{\textbf{Mean difference}\\\textbf{(percentage points)}} & \textbf{95\% CI} & \textbf{Wilcoxon $p$} \\
\midrule
SE & -1.85 & $[-7.17,\ 1.72]$ & 0.974 \\
Synthetic & -0.34 & $[-0.80,\ 0.03]$ & 0.356 \\
Pressure & 0.70 & $[-0.22,\ 1.74]$ & 0.106 \\
\bottomrule
\end{tabular}
\end{table}

\subsection{Extended error and reliability metrics}
Table~\ref{tab:extended_metrics} complements the primary capped-MPE results with uncapped MPE, raw engineering-value MAE/RMSE, full-scale tail errors and operational tolerance pass rates. Across the nine all-domain and LODO evaluations, all 615 generated answers were numerically parseable (0 parsing failures).

\begin{table}[!htbp]
\centering
\caption{Extended error, tail-error and operational tolerance metrics for fine-tuned Qwen2.5-VL-7B models. MAE and RMSE use dataset-specific engineering units; P95 FS and Max FS indicate 95th-percentile and maximum errors relative to full scale (FS). Pass rates denote the percentage of predictions within operational tolerances ($\pm1\%$, $\pm2\%$, $\pm5\%$ FS). No parsing failures occurred across evaluated samples (0/205 per model condition).}
\label{tab:extended_metrics}
\footnotesize 
\setlength{\tabcolsep}{2pt} 
\begin{tabular}{llcccccccc}
\toprule
\textbf{Dataset} & \textbf{Range} & \makecell{\textbf{Uncapped}\\\textbf{MPE (\%)}} & \textbf{MAE} & \textbf{RMSE} & \makecell{\textbf{P95 FS}\\\textbf{(\%)}} & \makecell{\textbf{Max FS}\\\textbf{(\%)}} & \makecell{\textbf{Pass}\\\textbf{$\pm1\%$}} & \makecell{\textbf{Pass}\\\textbf{$\pm2\%$}} & \makecell{\textbf{Pass}\\\textbf{$\pm5\%$}} \\
\midrule
SE & with & 4.43 & 48.39 & 176.47 & 19.90 & 41.25 & 44.2\% & 60.5\% & 79.1\% \\
Synthetic & with & 2.39 & 0.17 & 0.61 & 5.00 & 80.00 & 37.3\% & 64.3\% & 96.0\% \\
Pressure & with & 3.31 & 0.13 & 0.17 & 9.38 & 16.40 & 19.4\% & 36.1\% & 83.3\% \\
\midrule
SE & without & 13.42 & 62.25 & 278.93 & 26.44 & 407.14 & 46.5\% & 55.8\% & 81.4\% \\
Synthetic & without & 5.94 & 0.45 & 3.61 & 9.06 & 505.00 & 37.3\% & 64.3\% & 92.9\% \\
Pressure & without & 2.61 & 0.10 & 0.16 & 8.53 & 16.50 & 38.9\% & 61.1\% & 91.7\% \\
\bottomrule
\end{tabular}
\end{table}

The no-range SE and Synthetic results illustrate why uncapped reporting matters: their capped MPEs are 6.28\% and 2.73\% (Table~\ref{tab:final_comparison}), but uncapped MPE rises to 13.42\% and 5.94\%, respectively, because of extreme outliers. Dataset-level tail statistics therefore reveal failures that capped MPE alone can conceal.

\subsection{Leave-one-dataset-out transfer}
\label{subsec:lodo}
The LODO experiment directly tests whether performance survives removal of the target domain from fine-tuning. Table~\ref{tab:lodo} compares each held-out-domain model with the all-domain with-range model on the same test images.

\begin{table}[!htbp]
\centering
\caption{Leave-one-dataset-out (LODO) cross-domain transfer evaluation for fine-tuned Qwen2.5-VL-7B. LODO MPE is the range-normalized mean percentage error on the held-out test partition with its 95\% nonparametric bootstrap confidence interval [CI]. ``Increase (pp)'' represents the performance degradation in percentage points (LODO MPE minus the all-domain with-range baseline from Table~\ref{tab:final_comparison}), where positive values indicate increased error from domain exclusion. Wilcoxon $p$ is the two-tailed significance value from a paired Wilcoxon signed-rank test on per-image errors.}
\label{tab:lodo}
\small
\begin{tabular}{lcccc}
\toprule
\makecell{\textbf{Held-out}\\\textbf{dataset}} & \makecell{\textbf{LODO MPE}\\\textbf{[95\% CI]}} & \makecell{\textbf{Increase}\\\textbf{(pp)}} & \makecell{\textbf{Increase}\\\textbf{95\% CI}} & \makecell{\textbf{Wilcoxon}\\\textbf{$p$}} \\
\midrule
SE & 13.25 [8.38, 19.00] & 8.82 & [4.07, 14.20] & $<0.001$ \\
Synthetic & 14.99 [11.14, 19.25] & 12.60 & [8.37, 17.21] & $<0.001$ \\
Pressure & 4.99 [3.09, 7.26] & 1.68 & [-0.06, 3.93] & 0.360 \\
\bottomrule
\end{tabular}
\end{table}

Excluding SE or Synthetic from training causes a large and statistically significant increase in error, while the Pressure Gauge degradation is not significant in the present 36-image test set. For Pressure Gauge, the reported 1.68-percentage-point increase is relative to the all-domain \emph{with-range} baseline of 3.31\%, not the 2.61\% no-range result that is the best all-domain result for that dataset. These results constrain the generalization claim: the specialized model transfers imperfectly across domains and benefits materially from exposure to visually related training data.

\subsection{Adverse-condition robustness}
\label{subsec:robustness}
\begin{figure}[!htbp]
\centering
\includegraphics[width=0.82\linewidth]{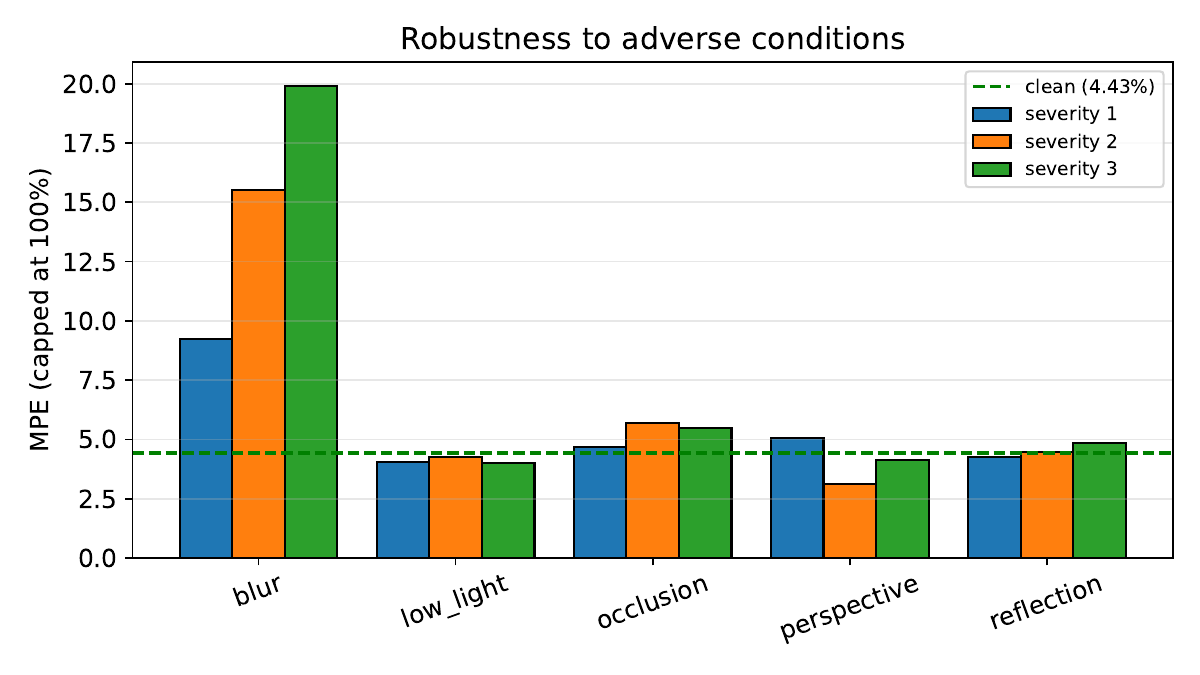}
\caption{MPE under synthetic adverse-condition corruptions on the SE test set. Blur is the dominant tested degradation.}
\label{fig:robustness_mpe}
\end{figure}

Robustness was evaluated on all 43 SE test images by applying deterministic synthetic corruptions at three severity levels. Blur uses Gaussian radii 1.5, 3.0 and 5.0 pixels; low light multiplies brightness by 0.6, 0.4 and 0.25; reflection overlays a blurred bright ellipse with increasing opacity; perspective perturbation increases corner displacement; and occlusion adds one, two or three small randomly positioned black patches without targeting any semantic region. The clean with-range SE baseline is 4.43\% MPE. Table~\ref{tab:robustness} reports the corruption results numerically, while Figure~\ref{fig:robustness_mpe} shows the MPE trend across severity levels.

\begin{table}[!htbp]
\centering
\caption{SE corruption robustness for the fine-tuned with-range Qwen2.5-VL-7B model. Values are capped MPE (\%). The final column is the severity-3 failure rate above 5\% full scale.}
\label{tab:robustness}
\small
\begin{tabular}{lcccc}
\toprule
\textbf{Corruption} & \makecell{\textbf{Severity}\\\textbf{1}} & \makecell{\textbf{Severity}\\\textbf{2}} & \makecell{\textbf{Severity}\\\textbf{3}} & \makecell{\textbf{Fail $>5\%$}\\\textbf{at S3}} \\
\midrule
Gaussian blur & 9.24 & 15.51 & 19.92 & 65.12\% \\
Low light & 4.05 & 4.27 & 4.00 & 20.93\% \\
Occlusion & 4.68 & 5.69 & 5.47 & 23.26\% \\
Perspective & 5.07 & 3.11 & 4.14 & 16.28\% \\
Reflection & 4.26 & 4.47 & 4.86 & 23.26\% \\
\bottomrule
\end{tabular}
\end{table}

The model is notably sensitive to blur, whereas the tested low-light, reflection, perspective and small-occlusion perturbations remain closer to the clean baseline. The strong blur degradation persists despite the moderate Gaussian-blur augmentation used during training, indicating that generic blur augmentation alone does not provide adequate robustness at the tested severe blur levels. These tests increase controlled coverage of adverse conditions but should not be interpreted as a substitute for a larger real-world environmental benchmark.

\subsection{Visual endpoint probe}
Using the scale probe prompt defined in Table~\ref{tab:prompts}, the visual endpoint experiment evaluates whether the fine-tuned model can extract minimum and maximum scale numbers directly from the dial image without supplying them in the prompt. Across 43 SE test images, the minimum endpoint is parsed on 100\% of images with 95.3\% exact-match accuracy and MAE 2.09, while the maximum endpoint has 100\% parse coverage, 83.7\% exact-match accuracy and MAE 44.21. This supports the narrow conclusion that the VLM often reads printed scale endpoints visually; it does \emph{not} demonstrate reliable understanding of tick intervals, units or every scale label. Range metadata therefore remains useful in applications where a gauge registry or trusted configuration is available.

\subsection{Calibration and high-confidence errors}

\begin{figure}[!htbp]
\centering
\begin{subfigure}[b]{0.49\textwidth}
\centering
\includegraphics[width=\linewidth]{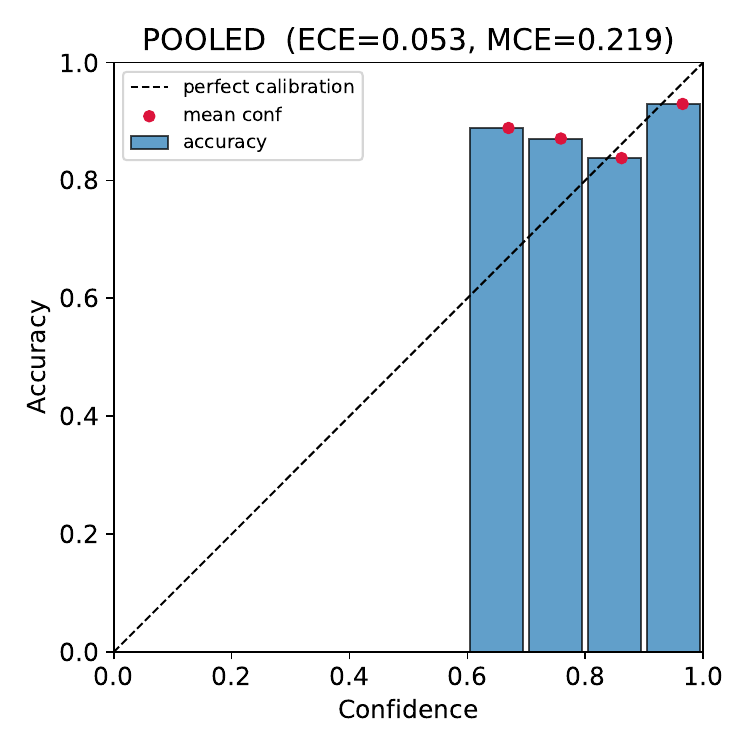}
\caption{Pooled reliability diagram.}
\end{subfigure}\hfill
\begin{subfigure}[b]{0.48\textwidth}
\centering
\includegraphics[width=\linewidth]{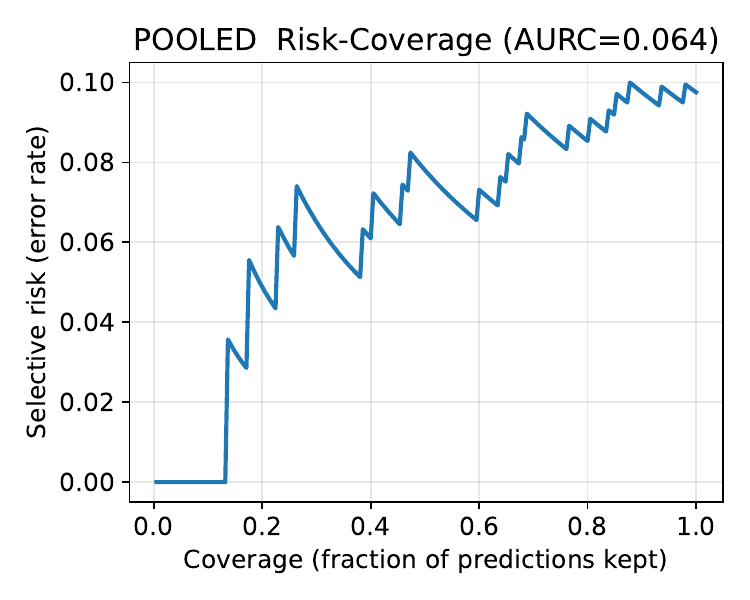}
\caption{Pooled risk-coverage curve.}
\end{subfigure}
\caption{Calibration analysis for the final with-range model. Corresponding no-range plots will be included in the public result package described in Section~\ref{sec:data_code}.}
\label{fig:calibration}
\end{figure}

Table~\ref{tab:calibration} reports calibration metrics evaluated under the formulation detailed in Section~\ref{sec:confidence_and_calibration}. Across 205 pooled test predictions, the with-range model has ECE 0.053 and AURC 0.064, with two incorrect predictions at sequence confidence $\ge0.99$. The no-range model has ECE 0.047 and AURC 0.055, with one such high-confidence error. Thus, confidence is informative enough to rank some predictions but is not a reliable stand-alone safety signal. Figure~\ref{fig:calibration} shows the pooled reliability diagram and risk-coverage curve for the with-range model.

\begin{table}[!htbp]
\centering
\caption{Confidence calibration and reliability metrics for the fine-tuned Qwen2.5-VL-7B models evaluated on individual datasets and the pooled test set ($N=205$). Accuracy@5\% denotes the proportion of predictions meeting the $\pm5\%$ full-scale operational correctness threshold. Expected Calibration Error (ECE) and Maximum Calibration Error (MCE) quantify average and worst-case calibration gaps across 10 equal-width confidence bins. AURC represents the Area Under the Risk-Coverage Curve measuring error rejection efficacy under confidence filtering. Mean Confidence denotes the average token-sequence confidence ($C$) and High-Confidence (HC) errors count critical failures where absolute error exceeds 5\% of full scale despite sequence confidence $C \ge 0.99$.}
\label{tab:calibration}
\small
\setlength{\tabcolsep}{2.5pt}
\begin{tabular}{llcccccc}
\toprule
\textbf{Model} & \textbf{Dataset} & \makecell{\textbf{Accuracy}\\\textbf{@5\% (\%)}} & \textbf{ECE} & \textbf{MCE} & \textbf{AURC} & \makecell{\textbf{Mean}\\\textbf{confidence}} & \makecell{\textbf{HC}\\\textbf{errors}} \\
\midrule
With range & SE & 79.1 & 0.158 & 0.311 & 0.177 & 0.920 & 1 \\
With range & Synthetic & 96.0 & 0.036 & 0.357 & 0.021 & 0.925 & 1 \\
With range & Pressure & 83.3 & 0.138 & 0.198 & 0.175 & 0.803 & 0 \\
With range & Pooled & 90.2 & 0.053 & 0.219 & 0.064 & 0.902 & 2 \\
\midrule
No range & SE & 81.4 & 0.131 & 0.353 & 0.066 & 0.945 & 0 \\
No range & Synthetic & 92.1 & 0.045 & 0.068 & 0.049 & 0.926 & 1 \\
No range & Pressure & 91.7 & 0.079 & 0.328 & 0.035 & 0.851 & 0 \\
No range & Pooled & 89.8 & 0.047 & 0.328 & 0.055 & 0.917 & 1 \\
\bottomrule
\end{tabular}
\end{table}

\subsection{Pressure Gauge video-level analysis and cross-study comparison}

The Pressure Gauge test set contains 36 images drawn from seven videos that are distinct from the training videos. As previously noted, our evaluation partition was deliberately curated to include only frames exhibiting distinct gauge readings, yielding a more diverse and arguably more demanding benchmark than the uniform, fixed-interval frame-sampling protocol used in the reference study \citep{HowellsEtAl2021CVPRW}. Because their purpose-built multi-stage mobile gauge-reading pipeline is not publicly available, their method cannot be re-executed directly on our specific 36-frame test split. Table~\ref{tab:howells} therefore combines the requested per-video breakdown with a transparent cross-study comparison using the relative-error metric $\mu_R$ defined in the reference study \citep{HowellsEtAl2021CVPRW}. The fine-tuned Qwen2.5-VL-7B metrics reflect the no-range condition, where the reported confidence intervals correspond to per-video bootstrap MPE intervals scaled to the relative-error domain (divided by 100).

\begin{table*}[!htbp]
\centering
\caption{Per-video Pressure Gauge comparison with the reference method \citep{HowellsEtAl2021CVPRW}. Values are mean relative error $\mu_R$ (lower is better). The Qwen column includes a 95\% bootstrap CI. Sampling protocols are not identical.}
\label{tab:howells}
\small
\begin{tabular}{lccc}
\toprule
\textbf{Video} & \textbf{$n$} & \textbf{Reference $\mu_R$} & \makecell{\textbf{Fine-tuned Qwen2.5-VL-7B}\\\textbf{$\mu_R$ [95\% CI]}} \\
\midrule
test1/man1 & 13 & 0.003 & 0.0096 [0.0058, 0.0138] \\
test2/man2 & 12 & 0.010 & 0.0319 [0.0137, 0.0596] \\
test3/man3 & 5 & 0.036 & 0.0320 [0.0240, 0.0408] \\
test4/man4 & 1 & 0.008 & 0.0080 [0.0080, 0.0080] \\
test5/man5 & 2 & No prediction & 0.0520 [0.0320, 0.0720] \\
test6/man6 & 1 & 0.045 & 0.1250 [0.1250, 0.1250] \\
test7/man7 & 2 & 0.042 & 0.0167 [0.0167, 0.0167] \\
\midrule
Overall & 36 & 0.024\textsuperscript{a} & 0.0261 [0.0166, 0.0380] \\
\bottomrule
\end{tabular}
\begin{flushleft}
\footnotesize \textsuperscript{a}The reference-study mean excludes test5, for which no prediction was produced.
\end{flushleft}
\end{table*}


The comparison is mixed rather than uniformly favorable. The reference method uses a task-specific multi-stage pipeline for the Pressure Gauge video setting, including localization, geometric processing and reading stages \citep{HowellsEtAl2021CVPRW}. Against that dataset-specific methodology, the general VLM-based reader has lower mean relative error on test3 and test7, matches test4, has higher error on test1, test2 and test6 and returns predictions for test5 where the multi-stage method reports no result. Its overall relative error is therefore similar despite using a direct cross-dataset reading formulation rather than a pipeline specifically designed for this dataset. Because the sampled frames are not identical, this is evidence of competitive performance and broader coverage, not a controlled claim of superiority.

\subsection{Other model baselines and computational cost}
The study also evaluates Qwen2.5-VL-3B, compact SmolVLM2 models and a MobileNetV2 regressor on SE. Table~\ref{tab:se_comparison} reports these additional baseline accuracies; the primary Qwen2.5-VL-7B SE results are reported once, in Table~\ref{tab:final_comparison}. Because per-image outputs for the additional baselines are not available under the same fixed-epoch Qwen2.5-VL-7B protocol, the table is used for architectural context rather than paired significance claims. The MobileNetV2 with-range and without-range experiments also use different target formulations (normalized position versus raw-value regression), so their difference is not interpreted as a clean metadata ablation.

\begin{table}[!htbp]
\centering
\caption{Additional model baselines on the SE dataset. MPE is reported in percent. The primary Qwen2.5-VL-7B values (4.43\% with range; 6.28\% without range) are reported in Table~\ref{tab:final_comparison} and are not duplicated here.}
\label{tab:se_comparison}
\small
\begin{tabular}{lcc}
\toprule
\multirow{2}{*}{\textbf{Model}} & \multicolumn{2}{c}{\textbf{MPE (\%)}} \\
\cmidrule(lr){2-3}
& \makecell{\textbf{With}\\\textbf{range}} & \makecell{\textbf{Without}\\\textbf{range}} \\
\midrule
Qwen2.5-VL-3B-Instruct & 4.98 & 7.33 \\
SmolVLM2-500M-Instruct & 24.14 & 21.12 \\
SmolVLM2-256M-Instruct & 26.22 & 17.78 \\
MobileNetV2 Regressor & 11.12 & 27.82 \\
\bottomrule
\end{tabular}
\end{table}

A protocol-matched efficiency benchmark was then run for all five architectures on the same Nvidia RTX A5500 using the same source image, batch size of one, five warm-up iterations, 30 timed trials, CUDA synchronization and peak-memory procedure, with preprocessing excluded from the timed region. VLM latency measures greedy generation with a maximum of 20 new tokens; MobileNetV2 latency measures a single forward pass. Table~\ref{tab:efficiency_benchmark} reports mean and 95th-percentile latency and peak allocated GPU memory during inference. Training duration is reported on a per-epoch basis from these later protocol-matched benchmark runs rather than extrapolated from the main accuracy-training logs or to a total multi-epoch training time. These timing measurements were collected after the main accuracy experiments and were not used for model or checkpoint selection. Peak GPU memory during training was not recorded in this benchmark; the memory values in Table~\ref{tab:efficiency_benchmark} are inference-only peak allocations.

\begin{table*}[!htbp]
\centering
\caption{Protocol-matched computational benchmark on an Nvidia RTX A5500. Peak memory is peak allocated GPU memory during inference; training time is measured per epoch under the benchmark configuration.}
\label{tab:efficiency_benchmark}
\footnotesize
\setlength{\tabcolsep}{1.5pt}
\begin{tabular}{@{}lrrrr@{}}
\toprule
\textbf{Model} & \makecell{\textbf{Mean latency}\\\textbf{(ms)}} & \makecell{\textbf{P95 latency}\\\textbf{(ms)}} & \makecell{\textbf{Peak GPU}\\\textbf{memory (MB)}} & \makecell{\textbf{Training time}\\\textbf{per epoch (min)}} \\
\midrule
Qwen2.5-VL-7B & 594.98 & 597.30 & 8101.53 & 8.05 \\
Qwen2.5-VL-3B & 599.74 & 604.12 & 6668.66 & 7.65 \\
SmolVLM2-500M & 582.55 & 587.18 & 2882.85 & 16.81 \\
SmolVLM2-256M & 536.53 & 539.52 & 1132.43 & 16.17 \\
MobileNetV2 & 3.95 & 4.07 & 143.73 & 0.65 \\
\bottomrule
\end{tabular}
\end{table*}

The VLMs have broadly similar sub-second generation latency under this protocol (approximately 537--600 ms), whereas MobileNetV2 is much faster at 3.95 ms. Peak inference memory decreases more consistently with model size, from 8101.53 MB for Qwen2.5-VL-7B to 1132.43 MB for SmolVLM2-256M; MobileNetV2 uses 143.73 MB. The near-equality of Qwen2.5-VL-7B and 3B latency illustrates that short autoregressive generation latency is not determined by parameter count alone. Likewise, the measured per-epoch training durations depend on implementation and processor overhead and do not decrease monotonically with model size. These timings exclude image acquisition, multi-gauge detection or instance selection, preprocessing, output parsing, network/database writes and alarm handling.

\subsection{Interpretability: blurred-patch sensitivity and attention rollout}
\label{sec:interpretability}

To examine which image regions influence the fine-tuned model's numerical predictions, we compare two complementary spatial diagnostics: blurred-patch perturbation and attention rollout. The blurred-patch analysis divides each image into a $10\times10$ grid and locally blurs one cell at a time. The resulting sensitivity map reflects the change in prediction error relative to the unperturbed image, thereby identifying regions for which loss of local visual detail most affects the generated reading. Unlike hard occlusion with a black patch, local blurring preserves the approximate image structure and reduces the possibility that the explanation is dominated by an artificial occlusion boundary.

Attention rollout provides a second, non-perturbative diagnostic. The rollout is computed from the final four transformer layers, beginning from positions associated with the generated non-EOS answer tokens and propagating mean-head self-attention, with residual normalization, back to the visual-token representation. To make eager-attention extraction computationally tractable, this diagnostic uses an aspect-ratio-preserving 448-pixel short side and a processor cap of 262,144 pixels. The resulting attention map is projected back to the source-image aspect ratio for visualization. Because this differs from the standard evaluation resolution, the attention maps are used only as qualitative spatial diagnostics; all numerical reading results are obtained from the standard evaluation pipeline.

\begin{figure*}[!htbp]
\centering
\newcommand{\subfigwidth}{0.23\textwidth} 

\begin{subfigure}[c]{\subfigwidth}\centering
\includegraphics[width=\linewidth]{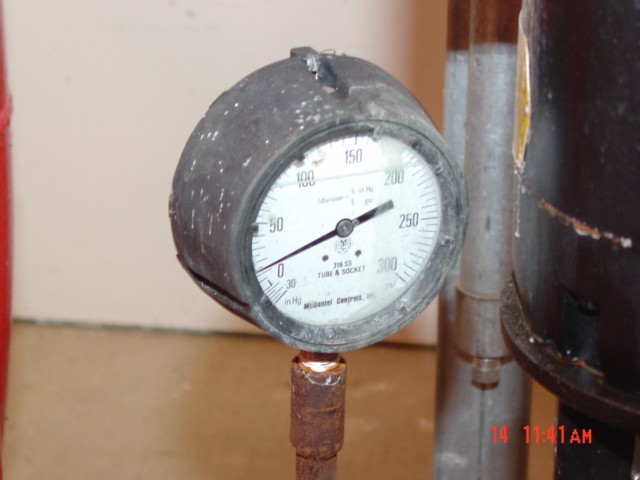}\caption{Perspective input}\end{subfigure}\hfill
\begin{subfigure}[c]{\subfigwidth}\centering
\includegraphics[width=\linewidth]{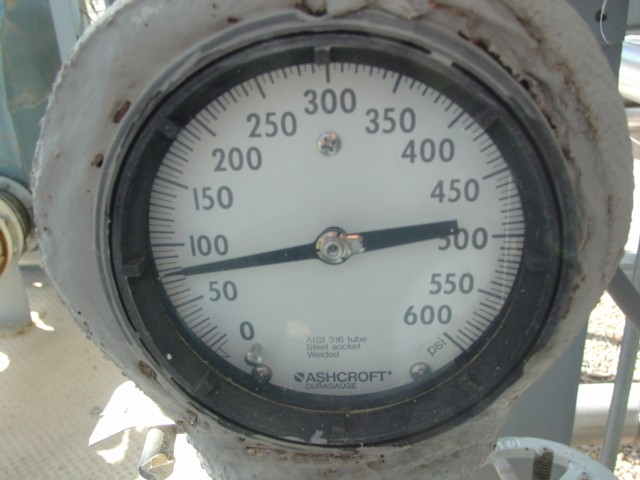}\caption{Dense-scale input}\end{subfigure}\hfill
\begin{subfigure}[c]{\subfigwidth}\centering
\includegraphics[width=\linewidth]{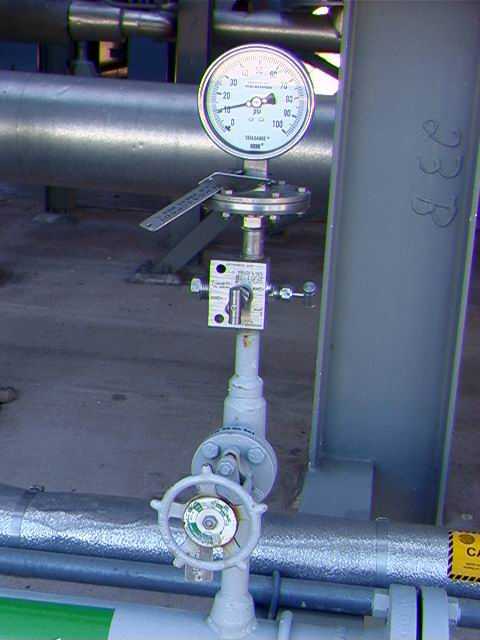}\caption{Anomalous input}\end{subfigure}

\vspace{2mm}
\begin{subfigure}[c]{\subfigwidth}\centering
\includegraphics[width=\linewidth]{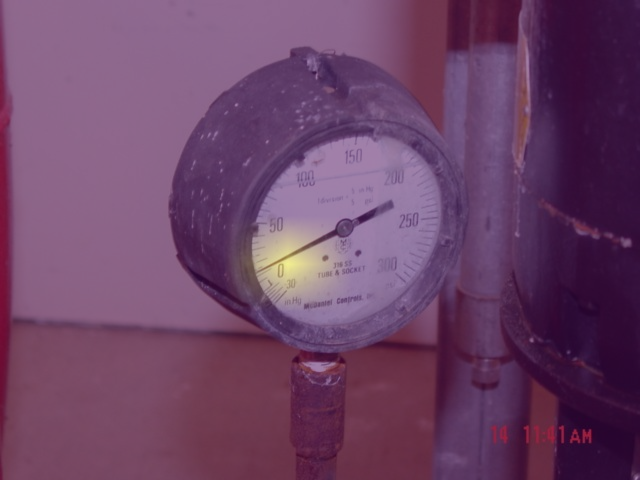}\caption{Blurred-patch sensitivity}\end{subfigure}\hfill
\begin{subfigure}[c]{\subfigwidth}\centering
\includegraphics[width=\linewidth]{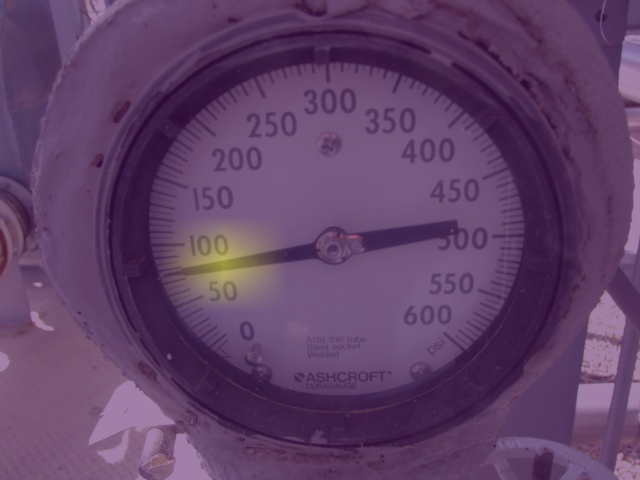}\caption{Blurred-patch sensitivity}\end{subfigure}\hfill
\begin{subfigure}[c]{\subfigwidth}\centering
\includegraphics[width=\linewidth]{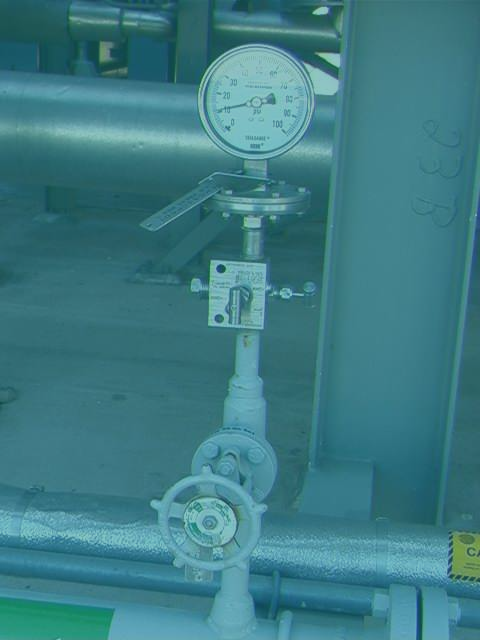}\caption{Blurred-patch sensitivity}\end{subfigure}

\vspace{2mm}
\begin{subfigure}[c]{\subfigwidth}\centering
\includegraphics[width=\linewidth]{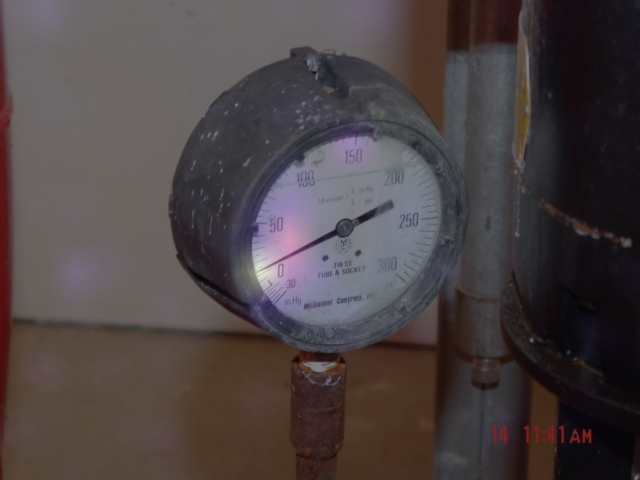}\caption{Attention rollout}\end{subfigure}\hfill
\begin{subfigure}[c]{\subfigwidth}\centering
\includegraphics[width=\linewidth]{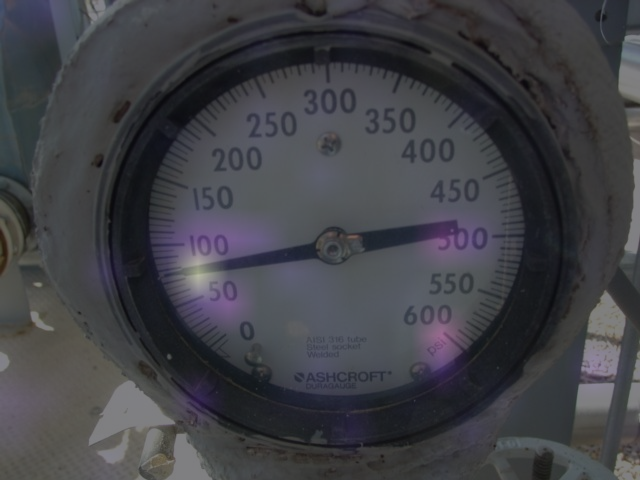}\caption{Attention rollout}\end{subfigure}\hfill
\begin{subfigure}[c]{\subfigwidth}\centering
\includegraphics[width=\linewidth]{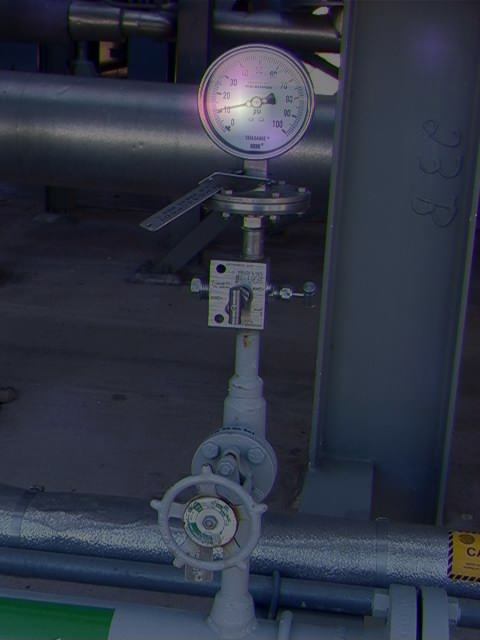}\caption{Attention rollout}\end{subfigure}

\caption{Interpretability analysis for three representative SE examples. Top: original inputs. Middle: blurred-patch sensitivity maps. Bottom: attention-rollout maps. The two successful examples emphasize similar gauge-face regions, though attention rollout is more spatially diffuse. The anomalous example shows a nearly flat sensitivity map despite attention on the gauge face, indicating that attention alone does not guarantee a correct numerical reading.}
\label{fig:interpretability_extended}
\end{figure*}

Figure~\ref{fig:interpretability_extended} presents three representative cases. Under the standard with-range evaluation, the perspective case has an actual reading of 12.5 and is predicted as 15.0 (0.76\% MPE), while the dense-scale case has an actual reading of 75 and is predicted as 70.0 (0.83\% MPE). In both successful examples, the blurred-patch maps concentrate their strongest sensitivity within the gauge face rather than the surrounding background. For the perspective case, the most influential region lies in the lower-left portion of the dial, close to the portion of the pointer and scale involved in determining the reading. The attention-rollout map is less spatially concentrated, but also assigns substantial attention to regions of the dial face. For the dense-scale case, blurred-patch sensitivity is again concentrated on the left side of the scale near the visually relevant dial markings, whereas the attention rollout is distributed across several locations on the gauge face, including both left- and right-side scale regions. Thus, the two explanation methods provide partially convergent rather than identical spatial evidence.

Quantitatively, the blurred-patch and attention-rollout maps have Pearson correlations of 0.775 for the perspective example and 0.659 for the dense-scale example and their dominant grid cell agrees in both cases. These values indicate moderate spatial agreement, while the visibly broader attention distributions caution against interpreting attention as a precise localization of the needle or a specific tick mark. In particular, no formal claim is made that either method identifies the exact pointer tip or corresponding scale label because spatial needle and tick-level annotations are not available for these examples.

The anomalous example illustrates the limitation of such post-hoc explanations. The image is a comparatively clear, approximately head-on industrial scene for which the actual reading is 10, yet the model predicts 41.0, corresponding to 31.0\% MPE, with sequence confidence 0.985. Its blurred-patch sensitivity map is effectively flat across the image: locally removing fine detail from any single grid cell produces no distinguishable sensitivity peak. Consequently, no meaningful perturbation-based localization can be inferred for this failure. In contrast, the attention-rollout map still places visible attention on the gauge face, particularly around the upper portion of the dial. This discrepancy is important: attention to the gauge does not establish that the model extracted the correct pointer geometry or numerical scale relationship. Because the perturbation-map variance is zero, a correlation between the two explanation maps is undefined for this example.

Taken together, the examples support a deliberately limited interpretation. For two accurate predictions, blurred-patch sensitivity and attention rollout show moderate spatial agreement and both emphasize regions of the gauge face, although attention is comparatively diffuse. For the anomalous prediction, attention remains concentrated on the gauge while perturbation sensitivity provides no meaningful localization. The analysis therefore does not support a causal claim that the erroneous numerical prediction arises specifically from the vision encoder, scale interpretation, or autoregressive decoding stage. Rather, it shows that apparently relevant attention can coexist with a large numerical error, reinforcing the need to treat post-hoc attention visualizations as diagnostic evidence rather than as proof of correct visual reasoning.

\section{Discussion: Industrial Risk Controls}
\label{sec:risk_controls}
The experiments expose several practical failure mechanisms that should shape deployment design. First, blur causes the largest tested corruption degradation, motivating an image-quality gate that rejects insufficiently focused inputs before numerical interpretation. Second, the nonzero high-confidence error rate shows that autoregressive token confidence must not be used as a sole validity signal. Third, LODO degradation indicates that a gauge reader should be validated on the target site or gauge family rather than assumed to transfer automatically.

A practical safety architecture could combine: (i) confidence-based abstention calibrated on the target environment; (ii) deterministic plausibility checks against the configured scale range and known process bounds; (iii) temporal consistency checks for video streams; (iv) an image-quality or blur detector; (v) redundant verification by OCR/geometric reasoning or a second model when a reading would trigger an alarm; and (vi) human confirmation for safety-critical decisions. These mechanisms are proposed safeguards, not components validated by the present experiments and should be evaluated before operational use.

\section{Limitations}
\label{sec:limitations}
Several limitations bound the conclusions of this study.
\begin{itemize}
    \item \textbf{Dataset size.} The test sets contain 126 Synthetic, 36 Pressure Gauge and 43 SE images. Bootstrap intervals quantify sampling uncertainty but do not replace a larger independent industrial benchmark.
    \item \textbf{Split structure.} Pressure Gauge is video-separated, but Synthetic and SE use random splits. The SE partition is not gauge-family- or plant-held-out, so related gauges or environments may occur in both training and test data. LODO measures transfer when the entire SE domain is excluded from training, but it does not identify gauge-type overlap within the random SE split.
    \item \textbf{Single-target-gauge input assumption.} Each input contains one target gauge, but images are not necessarily tight crops and may retain substantial scene context. Only multi-gauge source images were cropped to isolate individual instances. Automatic detection and selection among multiple gauges are not evaluated.
    \item \textbf{Range metadata.} The model can often read scale endpoints visually, but maximum-endpoint exact match is 83.7\% and tick intervals/units were not directly tested. A trusted range registry remains useful.
    \item \textbf{Cross-domain transfer.} LODO performance degrades substantially on Synthetic and SE, showing that the present model is not domain invariant.
    \item \textbf{Reliability.} High-confidence errors remain possible and blur sensitivity is substantial. Calibration and abstention must be validated for each deployment environment.

\end{itemize}


\section{Conclusion}
\label{sec:conclusion}
This study demonstrates that QLoRA specialization can substantially improve direct numerical reading by a general-purpose VLM for images containing a single target analog gauge, including images in which the gauge occupies only a small portion of the frame. Under a fixed 20-epoch protocol using the final checkpoint, the best MPEs are 2.39\% on Synthetic, 2.61\% on Pressure Gauge and 4.43\% on SE. Separate range/no-range models remove prompt-setting confounds, while expanded metrics expose tail errors that capped MPE alone would conceal.

The additional experiments also make the limits of the approach clear. Leave-one-dataset-out evaluation shows large transfer degradation on held-out Synthetic and SE; Gaussian blur is a strong failure mode; and high-confidence errors persist despite generally low average error. The evidence therefore supports a promising specialized single-target-gauge reader, not a deployment-ready system for detecting, selecting and reading multiple gauges throughout a plant. Future work should prioritize larger source-held-out industrial datasets, gauge-family-level evaluation, reliable scale/OCR extraction, calibrated abstention and integration with independent validation and multi-gauge detection.

\section{Acknowledgements}
The authors gratefully acknowledge the valuable guidance and support provided by Siemens Energy which was instrumental in the completion of this research. This research was also supported by The Florida High Tech Corridor’s Matching Grants Research Program (MGRP) under grant number GR109914.

\bibliographystyle{elsarticle-num}
\bibliography{references}

@article{ji2024another,
  title={Another way: Direct regression of meter readings for circular pointer meter images},
  author={Ji, Dongsheng and Zhang, Wenbo and Yang, Wen and Zhao, Qianchuan},
  journal={Engineering Applications of Artificial Intelligence},
  volume={136},
  pages={108863},
  year={2024},
  publisher={Elsevier}
}

@inproceedings{kappert2022sensor,
  title={Sensor systems for extremely harsh environments},
  author={Kappert, Holger and Schopferer, Sebastian and Saeidi, Nooshin and Ziesche, Steffen and Olowinsky, Alexander and Naumann, Falk and Jaegle, Martin and Spanier, Malte},
  booktitle={Sensors and Measuring Systems; 21th ITG/GMA-Symposium},
  pages={1--3},
  year={2022},
  organization={VDE}
}

@inproceedings{sandler2018mobilenetv2,
  title={Mobilenetv2: Inverted residuals and linear bottlenecks},
  author={Sandler, Mark and Howard, Andrew and Zhu, Menglong and Zhmoginov, Andrey and Chen, Liang-Chieh},
  booktitle={Proceedings of the IEEE conference on computer vision and pattern recognition},
  pages={4510--4520},
  year={2018}
}

@article{bordes2024introduction,
  title={An introduction to vision-language modeling},
  author={Bordes, Florian and Pang, Richard Yuanzhe and Ajay, Anurag and Li, Alexander C and Bardes, Adrien and Petryk, Suzanne and Ma{\~n}as, Oscar and Lin, Zhiqiu and Mahmoud, Anas and Jayaraman, Bargav and others},
  journal={arXiv preprint arXiv:2405.17247},
  year={2024}
}

@misc{grassme2022pressure,
  author       = {Julius Grassme},
  title        = {Pressure Gauge Reader Data},
  year         = {2022},
  howpublished = {\url{https://www.kaggle.com/datasets/juliusgrassme/pressure-gauge-reader-data}},
  note         = {Kaggle, Accessed: May 2025}
}

@article{LiEtAl2019Geometric,
  title={Automatic gauge detection via geometric fitting for safety inspection},
  author={Li, Beichen and Yang, Jingyu and Zeng, Xinyang and Yue, Huanjing and Xiang, Wei},
  journal={IEEE Access},
  volume={7},
  pages={87042--87048},
  year={2019},
  publisher={IEEE}
}

@article{Qwen25VL,
  title={Qwen2. 5-vl technical report},
  author={Bai, Shuai and Chen, Keqin and Liu, Xuejing and Wang, Jialin and Ge, Wenbin and Song, Sibo and Dang, Kai and Wang, Peng and Wang, Shijie and Tang, Jun and others},
  journal={arXiv preprint arXiv:2502.13923},
  year={2025}
}

@article{LoRA,
  title={Lora: Low-rank adaptation of large language models.},
  author={Hu, Edward J and Shen, Yelong and Wallis, Phillip and Allen-Zhu, Zeyuan and Li, Yuanzhi and Wang, Shean and Wang, Lu and Chen, Weizhu and others},
  journal={ICLR},
  volume={1},
  number={2},
  pages={3},
  year={2022}
}

@article{QLoRA,
  title={Qlora: Efficient finetuning of quantized llms},
  author={Dettmers, Tim and Pagnoni, Artidoro and Holtzman, Ari and Zettlemoyer, Luke},
  journal={Advances in neural information processing systems},
  volume={36},
  pages={10088--10115},
  year={2023}
}

@misc{EndavaKaggle,
  author = {Hanzelka, Jon and Tiday, Luke and Buinoschi-Tirpescu, Anda and Danciu, Dan},
  title = {Synthetic Data for Precision Gauge Reading},
  institution = {Endava},
  howpublished = {\url{https://www.kaggle.com/datasets/endava/synthetic-data-for-precision-gauge-reading/data}},
  note         = {Kaggle. Accessed: May 2025}
}

@article{YangEtAl2025Info,
  title={Automatic Reading Method for Analog Dial Gauges with Different Measurement Ranges in Outdoor Substation Scenarios},
  author={Yang, Yueping and Liao, Wenlong and Fan, Songhai and Hou, Jin and Tang, Hao},
  journal={Information},
  volume={16},
  number={3},
  pages={226},
  year={2025},
  publisher={MDPI}
}

@inproceedings{TrairattanapaEtAl2022Isai,
  title={Real-time multiple analog gauges reader for an autonomous robot application},
  author={Trairattanapa, Visarut and Phimsiri, Sasin and Utintu, Chaitat and Cherdchusakulcha, Riu and Tosawadi, Teepakorn and Thamwiwatthana, Ek and Tungjitnob, Suchat and Tangamonsiri, Peemapol and Takutruea, Aphisit and Keomeesuan, Apirat and others},
  booktitle={2022 17th International Joint Symposium on Artificial Intelligence and Natural Language Processing (iSAI-NLP)},
  pages={1--6},
  year={2022},
  organization={IEEE}
}

@inproceedings{ReitsmaEtAl2024Icra,
  title={Under pressure: learning-based analog gauge reading in the wild},
  author={Reitsma, Maurits and Keller, Julian and Blomqvist, Kenneth and Siegwart, Roland},
  booktitle={2024 IEEE International Conference on Robotics and Automation (ICRA)},
  pages={14--20},
  year={2024},
  organization={IEEE}
}

@inproceedings{LiEtAl2020JPhysConf,
  title={Analog gauge reader based on image recognition},
  author={Li, Xingxing and Yin, Panpan and Duan, Chao and Zhi, Yan},
  booktitle={Journal of Physics: Conference Series},
  volume={1650},
  number={3},
  pages={032061},
  year={2020},
  organization={IOP Publishing}
}

@inproceedings{PeixotoEtAl2022Telecom,
  title={Development of an analog gauge reading solution based on computer vision and deep learning for an iot application},
  author={Peixoto, Jo{\~a}o and Sousa, Jo{\~a}o and Carvalho, Ricardo and Santos, Gon{\c{c}}alo and Mendes, Joaquim and Cardoso, Ricardo and Reis, Ana},
  booktitle={Telecom},
  volume={3},
  number={4},
  pages={564--580},
  year={2022},
  organization={MDPI}
}

@article{NinamaEtAl2024SciRep,
  title={Computer vision and deep transfer learning for automatic gauge reading detection},
  author={Ninama, Hitesh and Raikwal, Jagdish and Ravuri, Ananda and Sukheja, Deepak and Bhoi, Sourav Kumar and Jhanjhi, NZ and Elnour, Asma Abbas Hassan and Abdelmaboud, Abdelzahir},
  journal={Scientific Reports},
  volume={14},
  number={1},
  pages={23019},
  year={2024},
  publisher={Nature Publishing Group UK London}
}

@inproceedings{LauridsenEtAl2019Visigrapp,
  title={Reading circular analogue gauges using digital image processing},
  author={Lauridsen, Jakob S and Graasm{\'e}, Julius AG and Pedersen, Malte and Jensen, David Getreuer and Andersen, S{\o}ren Holm and Moeslund, Thomas B},
  booktitle={14th International joint conference on computer vision, imaging and computer graphics theory and applications (Visigrapp 2019)},
  pages={373--382},
  year={2019},
  organization={SCITEPRESS Digital Library}
}

@inproceedings{TianEtAl2024Mipr,
  title={GaugeTracker: AI-Powered Cost-Effective Analog Gauge Monitoring System},
  author={Tian, Beitong and Wu, Mingyuan and Zhang, Ruixiao and Zheng, Haozhen and Chen, Bo and Wang, Yaohui and Trivedi, Shiv and Zhang, Shanbo and Kaufman, Robert Bruce and Espenhahn, Leah and others},
  booktitle={2024 IEEE 7th International Conference on Multimedia Information Processing and Retrieval (MIPR)},
  pages={477--483},
  year={2024},
  organization={IEEE}
}

@inproceedings{HowellsEtAl2021CVPRW,
  title={Real-time analogue gauge transcription on mobile phone},
  author={Howells, Ben and Charles, James and Cipolla, Roberto},
  booktitle={Proceedings of the IEEE/CVF conference on computer vision and pattern recognition},
  pages={2369--2377},
  year={2021}
}

@inproceedings{GellaboinaEtAl2013,
  title={Analog dial gauge reader for handheld devices},
  author={Gellaboina, Mahesh Kumar and Swaminathan, Gurumurthy and Venkoparao, Vijendran},
  booktitle={2013 IEEE 8th Conference on Industrial Electronics and Applications (ICIEA)},
  pages={1147--1150},
  year={2013},
  organization={IEEE}
}

@article{ZuoEtAl2020,
  title={A robust approach to reading recognition of pointer meters based on improved mask-RCNN},
  author={Zuo, Lin and He, Peilin and Zhang, Changhua and Zhang, Zhehan},
  journal={Neurocomputing},
  volume={388},
  pages={90--101},
  year={2020},
  publisher={Elsevier}
}

@article{LiuEtAl2020Measurement,
  title={A detection and recognition system of pointer meters in substations based on computer vision},
  author={Liu, Yang and Liu, Jun and Ke, Yichen},
  journal={Measurement},
  volume={152},
  pages={107333},
  year={2020},
  publisher={Elsevier}
}

\end{document}